\documentclass[10pt, a4paper]{article}
\usepackage{lrec}
\usepackage{multibib}
\newcites{languageresource}{Language Resources}
\usepackage{graphicx}
\usepackage{tabularx}
\usepackage{soul,subcaption}
\usepackage[toc,page]{appendix}
\usepackage{multicol,float}

\usepackage{epstopdf}
\usepackage[latin1]{inputenc}

\usepackage{hyperref}
\usepackage{xstring}

\title{The Spot the Difference corpus: a multi-modal corpus of spontaneous task oriented spoken interactions}

\name{Jos\'{e} Lopes, Nils Hemmingsson, Oliver \r{A}strand}

\address{KTH Royal Institute of Technology \\
         Stockholm, Sweden\\
         jdlopes@kth.se\\}

\abstract{
This paper describes the Spot the Difference Corpus which contains 54 interactions between pairs of subjects interacting to find differences in two very similar scenes. The setup used, the participants' metadata and details about collection are described. We are releasing this corpus of task-oriented spontaneous dialogues. This release includes rich transcriptions, annotations, audio and video. We believe that this dataset constitutes a valuable resource to study several dimensions of human communication that go from turn-taking to the study of referring expressions. In our preliminary analyses we have looked at task success (how many differences were found out of the total number of differences) and how it evolves over time. In addition we have looked at scene complexity provided by the RGB components' entropy and how it could relate to speech overlaps, interruptions and the expression of uncertainty. We found there is a tendency that more complex scenes have more competitive interruptions. \\ \newline  }

\begin{document}

\maketitleabstract

\section{Introduction}
\label{sec:Introduction}

Despite the recent advances in the field of Spoken Dialogue Systems (SDSs), non task-oriented spontaneous dialogue is still a very challenging problem since its structure is often difficult to represent, unlike task-oriented dialogues which could easily be represented by a flow chart. Therefore, DeVault \cite{devault2008contribution} compared task-oriented dialogue to assembling furniture. On the other hand, non task-oriented dialogue may be compared to dancing. To be able to dance one needs to learn the steps, but this might not be enough since dancing is a collaborative task where coordination is extremely important. Unlike what happens with the furniture assembling case, there is no going back to the point where the mistake was done. In spontaneous dialogues as in dancing the dialogue should continue, despite the mistakes. There are mechanisms that help to regulate these situations, for instance language, turn-taking and other types of non-verbal behavior. SDSs would greatly benefit if they could understand these mechanisms in order to be able to anticipate moments in a dialogue where some sort of communication breakdown is about to happen, so they could act in a more human like fashion both in realizing the breakdown and finding an appropriate solution to it. 

While replicating and studying these phenomena in a non task-oriented dialogue might be too complex given the current state-of-the-art of spoken dialogue systems, there are intermediate steps that can help us in this process. The step that we present in this paper is the Spot the Difference corpus. This is a corpus of task-oriented collaborative dialogues between humans using spontaneous speech. We think that this corpus is a valuable resource in the study of spontaneous dialogues both in terms of verbal and non-verbal behavior since the corpus release includes several annotations, audio and video data from the interactions. Unlike other similar tasks such as the Map Task \cite{anderson1991hcrc}, participants are free to choose the order in which they could discuss the objects in the scene. These resulted in a less structured data, but certainly richer in spontaneity.

In this paper we will describe the Spot the Difference corpus in detail: the experimental setup used, the whole procedure, the participants data and the annotation already performed on this data. To prove the usefulness of the corpus, we performed a preliminary analysis where we studied if certain aspects of communication are related to the complexity of the task. 

\section{Background}
\label{sec:background}
There are several mechanisms and efforts that can be used to improve human communication. In the particular case of dialogues occurring in the scope of a collaborative tasks, these efforts serve the purpose of achieving the common goal by the participants in the dialogue. 

These mechanisms or efforts can be expressed at the linguistic level. For instance, several studies have shown how important entrainment and coordination can relate to success in task oriented dialogues. Whereas entrainment presupposes an adaptation between speakers over time, coordination can be present from the very beginning of the dialogue in the way the speakers interact with each other. There is a fair number of studies which are focused on the role of lexical and syntactic coordination in both in task and non task-oriented dialogues. For instance, studies by \cite{GarrodAnderson87} and \cite{brennan1996conceptual} have focused on participants' coordination in terms of lexical items. \cite{reitter2007predicting} showed that for task solving in dialogue, lexical and syntactic repetition is a reliable predictor of task success given the first five minutes of task oriented dialogue. \cite{friedberg2012lexical} found a significant difference in  the performance of student engineering groups related to lexical entrainment. The high performing groups, increased their entrainment over time, whereas the low performing groups tended to decrease their lexical entrainment with time.  \cite{nenkova2008high} investigated entrainment in the use of the most commonly used words in the Switchboard \cite{godfrey1992switchboard} and the Columbia Games corpus \cite{benus2007prosody}, as well as its perceived naturalness, flow and task success. Their results indicate that entrainment in commonly used words, can be a predictor of the perceived naturalness of the dialogue and is significantly correlated with task success. The aforementioned efforts are ways of optimizing the dialogue. Other linguistic mechanisms are normally used when the dialogue falls below the optimization line, for instance repairs. These have been previously studied in the context of the Map Task \cite{colman2011distribution}, where it was shown that patterns are cross-person and cross-turn. Given the nature of the task, the mechanisms above memtioned often go together with the study of referring expressions (RE) and reference resolution (RR). The corpora presented in \cite{zarriess2016pentoref} include examples where REs are improved over time to achieve a common ground. This process may require participants to use self-repairs in their utterances, and adjust them with their partner over time for efficiency purposes. The fact that the corpus we are releasing is multi-modal could benefit an integrated approach to improve the understanding of dialogue utterances such as the one presented in \cite{kennington2013interpreting}.

But these mechanisms and efforts are not exclusively linguistic. \cite{nenkova2008high} found that higher degrees of entrainment are associated with more overlaps and fewer interruptions. \cite{oviatt2015spoken} investigated overlapped speech in groups of students trying to jointly solve math problems. They found that during the most productive phases of the interactions the amount of overlap was higher when compared to other phases of the problem solving. Moreover, they could also show that the domain experts differed in the kind of interruptions they made from non-domain experts. \cite{goldberg1990interrupting} stated that interruptions may be used to convey rapport in competitive settings and \cite{poesio2010completions} mentioned them as signs of coordination and alignment. However, in a similar set up to the one used in this study, \cite{bull1998analysis} found that the complexity and the lack of familiarity with the tasks could result in longer gaps. 

Uncertainty display is another mechanism that indicates that the dialogue might approaching a point where some recovery strategy might be needed. It has already been studied in the scope of tutoring dialogues \cite{liscombe2005detecting}, but also in spontaneous speech \cite{schrank2015automatic}. Although tutoring dialogues can be seen as a collaborative dialogue, there is no short term goal, and therefore we hypothesize that the display of uncertainty will be different that we observe in our corpus, thus reinforcing the importance of the resource that we are releasing.

\section{Data collection}
\label{sec:Data}

\subsection{Procedure}
\label{sec:procedure}
The Spot the Difference corpus was recorded during 2016 at KTH. Participants were recruited via mailing lists and word of mouth. Participants were required to speak English and had to fill in a small personality questionnaire before the experiment. This questionnaire included a small subset of the Big Five Inventory \cite{john2008paradigm} with the 8 questions used to place individual in the introvert/extrovert axis. 36 participants took part in the experiment. Participants were informed that they were participating in an experiment to investigate human dialogues in a collaborative setting. 

\begin{figure}[h]
  \centering
  \begin{subfigure}[b]{0.22\textwidth}
  \includegraphics[width=\textwidth]{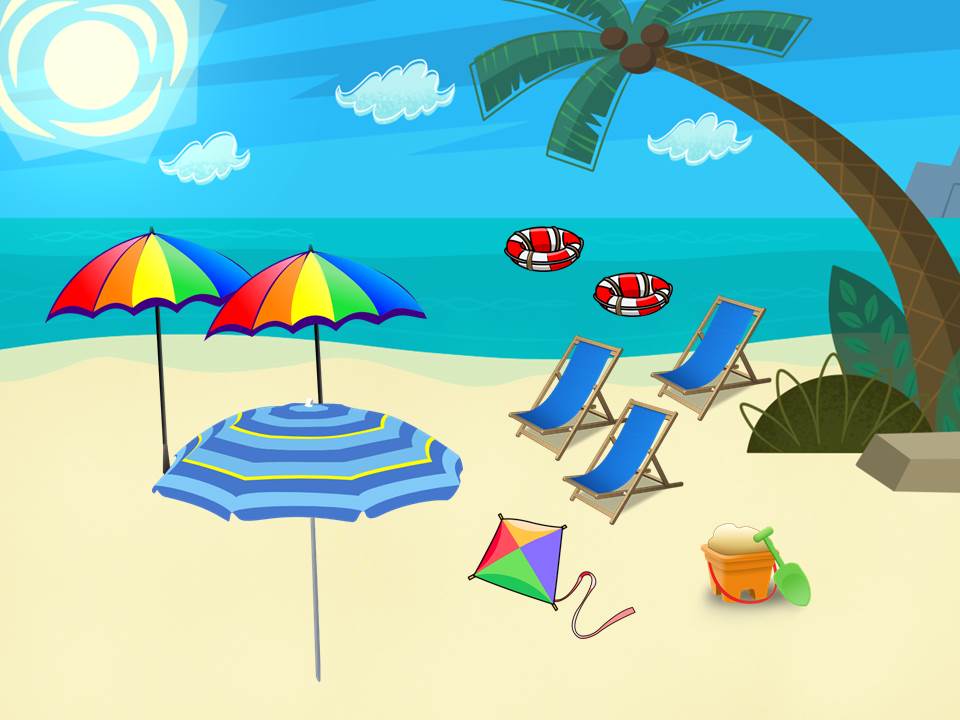}
  \end{subfigure}
  ~
  \begin{subfigure}[b]{0.22\textwidth}
  \includegraphics[width=\textwidth]{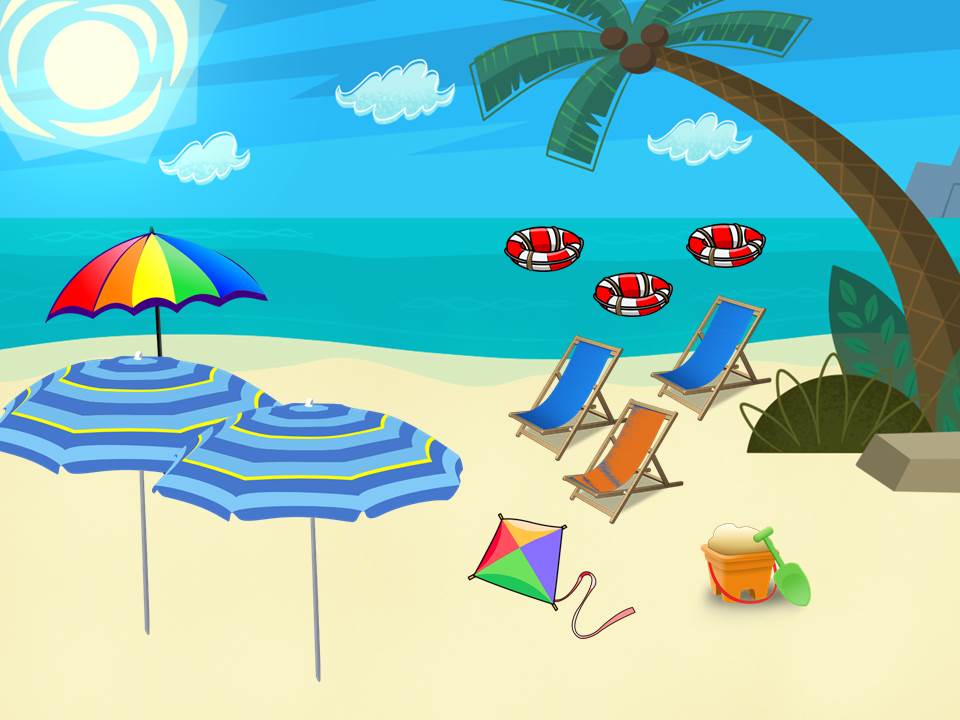}
  \end{subfigure}
  \caption{Example from the beach scenes used.}
  \label{fig:picdiff}
\end{figure}

They were briefly instructed about the task: they had to collaborate to find all the differences in two very similar scenes, such as those shown in Figure \ref{fig:picdiff}. Since they were sitting in different rooms with no visual contact, they were forced talk to each other in order to discuss the scene they had in front of them. To enable communication between the two of them, one of the head-mounted microphones used was connected to their partners computer speakers. They would describe the pictures in front of them and as they found differences, participants should engage in a sub-dialogue to locate where the difference was as precisely as possible and use the mouse to click in that area. They were informed that if they would not click the same area, the difference would not be recorded as found.  There were two roles in attributed in the beginning of the dialogue: the Instruction Giver (IG) and the other the Instruction Follower (IF). The IG had to lead the discussion by describing and locating the objects in the scene, whereas the IF had to follow to the IG's instructions, make clarification requests when necessary. An excerpt of a dialogue can be found in Table \ref{tab:dialogue}. The roles were randomly assigned once the participants did the first set of three scenes and kept over the course of the experiment. To get familiarized with the task participants had a chance to do a training scene before the experiment started. 

In \ref{tab:dialogue} we have transcribed a sample dialogue, where the participants are discussing the scene in Figure \ref{fig:picdiff}.

\begin{table}[htp]
\begin{center}
\begin{tabular}{l}
\textbf{IG}: OKAY THEN I HAVE LIKE $<$F-$>$ $<$\%aa$>$ \\THREE $<$\%aa$>$ $<$L-$>$ BEACH LYING CHAIRS\\
\textbf{IF}: OKAY\\
\textbf{IG}: YEAH\\
\textbf{IG}: YOU TOO?\\
\textbf{IF}: YES TWO OF THEM BLUE ONE IS ORANGE\\
\textbf{IF}: THAT\\
\textbf{IG}: OH THEY'RE $<$THE$>$ THE THREE BLUE\\
\textbf{IF}: OKAY THE ONE ORANGE IS THE ONE IN \\THE BOTTOM\\
\textbf{IG}: YEAH OKAY LET'S CLICK THERE I DON'T \\HAVE THAT\\
\end{tabular}
\end{center}
\caption{Excerpt of a dialogue transcriptions captured during the discussion about the scene pictured in Figure \ref{fig:picdiff}.}
\label{tab:dialogue}
\end{table}%

Whenever the time limit (200 seconds) was reached or the participants agreed to click the button to show the solution, the correct solution and the score was shown to both participants while the audio channel was kept open. Facing the solution the participants would have the possibility to discuss the missed differences and refine their strategy for the coming scenes. The IG had to click a button to continue to the following scene, once both partners had agreed to do so, unless it was the last scene of the set. In that case, no button would be shown. The procedure was repeated until each participant had completed three different sets. Including the necessary set ups, the experiment took about an hour. The participants would receive a cinema ticket as a compensation for their participation.

\subsection{Infrastructure}
The game was implemented with IrisTK \cite{Skantze2012}, already envisaging a future implementation in a dialogue system. IrisTK was running in parallel in two different machines, they were communicating with one another sending event messages, namely those generated by the eye-tracker and the mouse clicks on the scenes and respective coordinates. Each scene had a corresponding XML file with the description of the spatial arrangement of the objects in the scene, including the coordinates of the object, the radius, color, if the object corresponded to a difference, if the object was visible and possible ways of referring to that object.  

In the beginning of each set the streams were synchronized using a beep sound. This step was necessary since the timestamps for the IrisTK events (mouse clicks and eye tracking among others) were from the native computer.





\subsection{Participants}

From the 36 participants 14 were female subjects and 22 were male subjects. 9 of the female speakers were assigned the role of IG and 4 the role of IF. 9 of the male subjects were assigned the role of IG and the remaining 13 the role of IF. The average age of the participants was 34.3. Among the participants there were 18 different mother languages and only one native English speaker. All the non-native speakers but one were fluent in English and claimed to have English as their language at work. The most represented native language was Swedish (7 subjects), followed by Portuguese (6 subjects), Spanish and Farsi (4 participants) and French (2 participants). All the other native languages only had one participant. Only 4 dialogues out of the 54 were held between subjects with the same mother tongue. We avoided subjects that were acquainted from before to take part in the same session since this might have implications in their interaction. 



\subsection{Experimental setup}
Each set was composed of three scenes: an easy one, an average one and a difficult one. The difficulty level was assigned after the number of differences between pictures (more differences meaning higher difficulty). All the participants did each scene once, but the order of the scenes and the sets was randomized in order to avoid that the order of the scenes had an effect on the experiment. Since one of our goals was to create a corpus of spontaneous speech where we could study language, turn-taking and non-verbal behavior and how these evolve over time if partners don't change, we divided the subjects in two groups. The first group performed each set with a different partner (3 scenes with each partner). This will further on be referred as condition A. The second group performed all the 3 sets with the same partner, and this will be hereafter called condition B.

\subsection{Collected data}
The data recorded includes audio from the two head mounted microphones used by each participant. One of the microphones was the one used to communicate with the partner. From these microphones two mono audio files were generated per scene, one for each speaker. The other two microphones were connected to the same sound-card. From these microphones a stereo audio file was generated per set, with one speaker in each channel. Since these microphones recorded a complete set, the discussion that occurs once the solution is shown to the partipants until a new started is also included. 
We also recorded eye-tracking data with two gaming eye-trackers placed under each screen. The eye-tracking data contains both raw fixation data and fixation data for objects specified in the XML scene description. The raw fixation data was stored in the IrisTK log file. The fixation data for the objects was also saved in the IrisTK log file, but was, in addition, converted into a Praat Tier where each interval was labelled with the corresponding object identifier. Mouse click coordinates were also saved as a point tier in Praat. Finally, videos from two GoPro placed on top of each eye-trackers was also recorded for each set. 

\subsubsection{Transcription}

The audio for each speaker was recognized using the IBM Watson speech recognizer\footnote{https://www.ibm.com/watson/services/speech-to-text}. The output of the ASR was converted into a Praat interval tier. This allowed a reasonably accurate estimate of the content that was actually said, but most important, accurate time boundaries for the speech utterances in order to study turn taking behavior.

The data was also manually transcribed including disfluency annotations using the coding scheme defined in \cite{moniz2014speaking}. Part of this work was done on transcription made from scratch, while another part was done correcting the ASR output and inserting the disfluency annotation.


\subsubsection{Annotation}
Filled pauses were annotated separately by one annotator, without taking the transcription data into account. Annotated filled pauses were English filled pauses "eh, ah, aa, ahm" as well as filled pauses for the mother tongue of the participants. Since the transcription also coded Filled Pauses and was performed by a different annotator, it was possible to compute the agreement for Filled-Pause annotation which was in this case $0.61$, which can be seen as moderate agreement.

\begin{figure}[bp]
\begin{center}
\includegraphics[width=0.7\columnwidth]{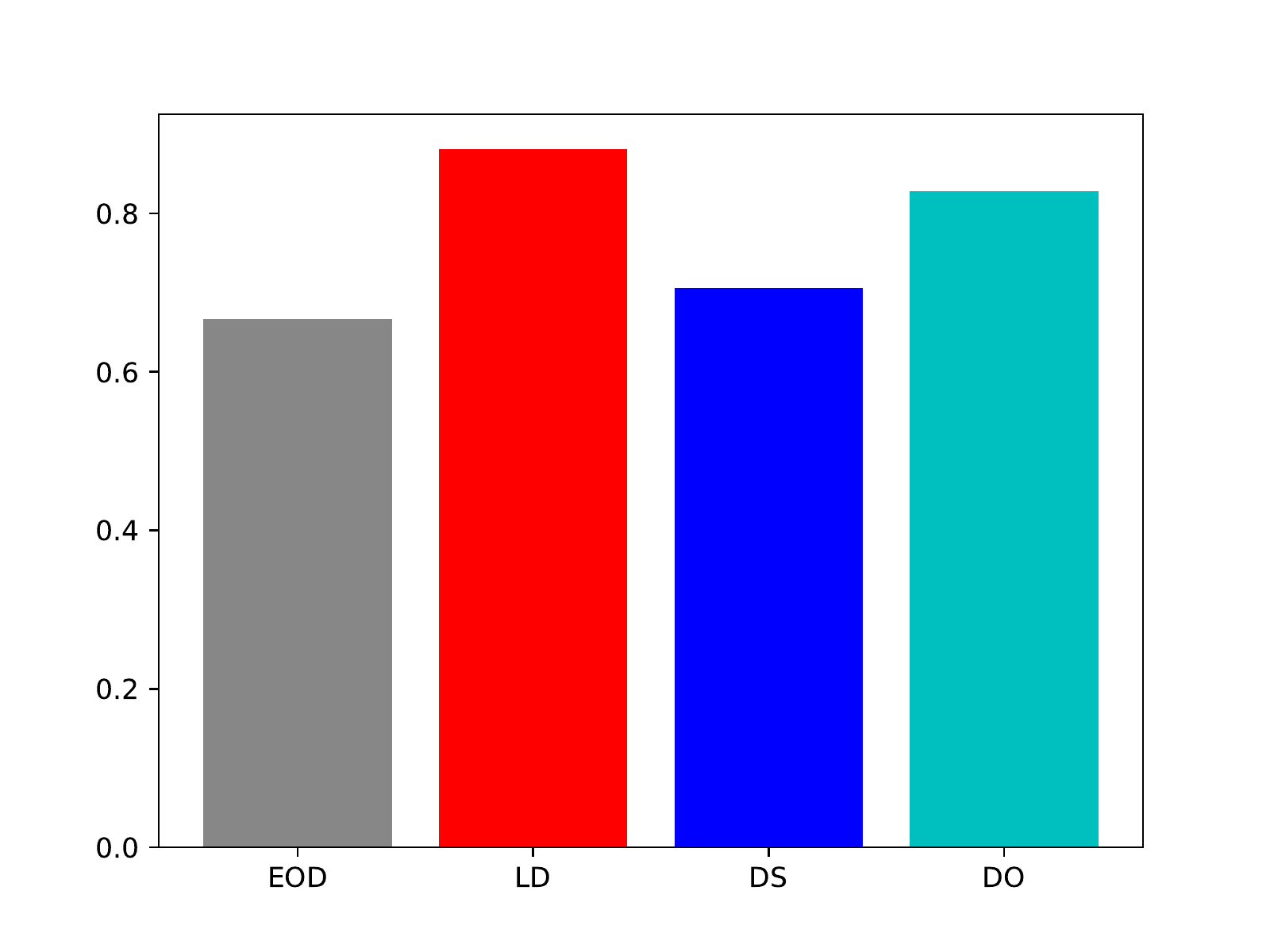}
\caption{Participation Equality per Topic, the closer to 1 the more even is the participation.}
\label{fig:ppeq}
\end{center}
\end{figure}

Each conversation was manually annotated per topic by one annotator. The topic labels used were Describing scene (DS), Describing object (DO), Locating difference (LD) and End of dialogue (EOD). DS corresponds to the segments where a participant is describing the scene focusing in the spatial relation between the objects in the scene. DO was used for segments where participants were describing some characteristic of a specific object. LD was used for the segments where participants were discussing the exact location of the difference. EOD dialogue corresponds to segments where participants are negotiating whether they should press the button to show the solution. These choice of topics followed an hypothesis that participants behaviour would change between topics as the graph in Figure \ref{fig:ppeq} shows for participation equality \cite{lai2013summarize}. 

We also automatically extracted all the speech overlaps found in the data and we annotated those which corresponded to floor changes as interruptions or not. Among overlaps, we considered interruptions whenever the interrupted speaker was not able complete the sentence. Therefore there might be other interruptions in the data that do not follow overlaps which, for the current analysis, were not taken into account. The interruptions were further divided into collaborative and competitive interruptions. We labeled as collaborative the interruptions those where the interrupting speaker completes the sentence and competitive interruptions were labeled whenever the interrupting speaker utters something unrelated to the interrupted sentence.

Furthermore, the video data was annotated for uncertainty using ELAN. In doing so, we formulated the binary definition "A conversation participant is \textit{uncertain} when they feel they do not understand what the counterpart is trying to communicate or that they do not know what to say". Those intervals in the videos where the annotator had this perception were annotated as uncertain and the remainder as certain. Each video was annotated by one annotator.

\subsubsection{Data formats}

Information about the dataset will be released in a JSON file. This JSON files contains all the dialogue ids and a link to a JSON set file. This JSON file includes metadata about the participants can be found, together with other relevant data about the session such as the scenes used, the data files (audio and video) and respective offsets, the log files and the annotations files for the scene: the Praat annotation file and the respective tiers for each participants and the uncertainty annotation ELAN file. Whenever the audio for the whole set was available, all the turns in the set, including those where participants are refining their strategy between scenes are included, with information about time boundaries, speaker, turn index, topic and rich transcriptions. The set JSON files further link to scene files. The scene JSON include information about the dialogue success, duration of the dialogue, scene audio files, and turns corresponding to the scene. In both set and scene JSON files, there is information about overlaps. The overlaps contain the time boundaries of the overlap, the speaker before and after the overlap, and the turn-index of the overlapping turn. The dataset can be accessed in \url{https://github.com/zedavid/SpotTheDifferenceData}.

\section{Data analysis}

Considering a dialogue as set of three scenes, 54 dialogues were recorded with this setup. Due to technical issues 4 of these dialogues could not be used in our data analysis. For this data we found that for each scene the average duration time was 188.3 seconds, meaning that there were scenes where the participants decided to unveil the solution before the time limit was reached. The average number of turns per dialogue was 121.4 (standard deviation 44.4).

Despite the fact that participants were assigned a specific role, in some dialogues we even observed these somehow reversed during the course of the interaction. In addition, as we mentioned before the participants could discuss their performance after the scene and refined their strategy, which can be seen as engaging factor and an incentive to improve the way they collaborate in the coming scenes.

\begin{figure}[htbp]
\begin{center}
\includegraphics[width=0.7\columnwidth]{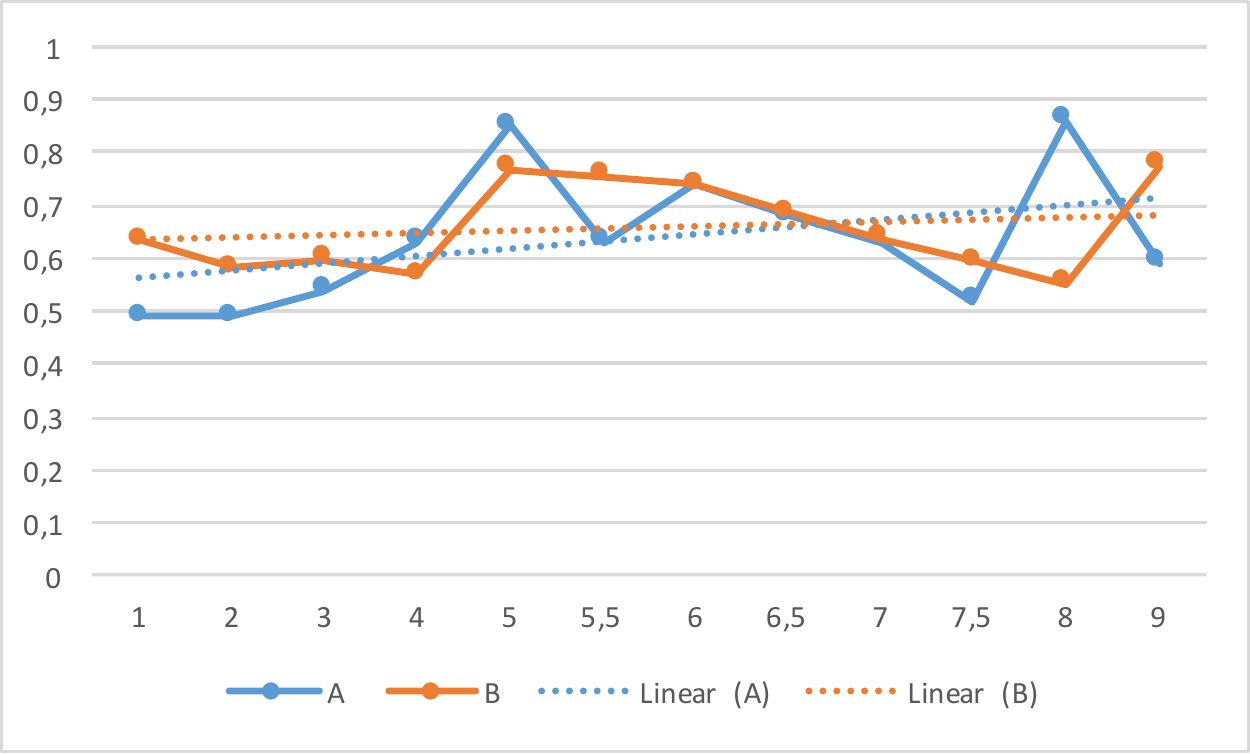}
\caption{Average relative number of differences found (y-axis) against average expertise in the task between the participants (x-axis).}
\label{fig:taskSuccessExperience}
\end{center}
\end{figure}

From the experimental design, we implicitly hypothesized that there was a learning curve both to get used to the task and to interact with each partner. If the latter would have played a significant role, we would expect that the participants in condition B would, as they accumulate experience, perform better that those in condition A. As a matter of fact, as seen in Figure \ref{fig:taskSuccessExperience}, we see that there is a similar progression in the task success achieved in each condition. According to the linear approximation, participants in condition A have improved more their task success than those in condition B during the course of the experiment. This means that the hypothesis there is a learning curve associated with the partner does not stand in our data.

\begin{figure}[htbp]
\begin{center}
\includegraphics[width=0.7\columnwidth]{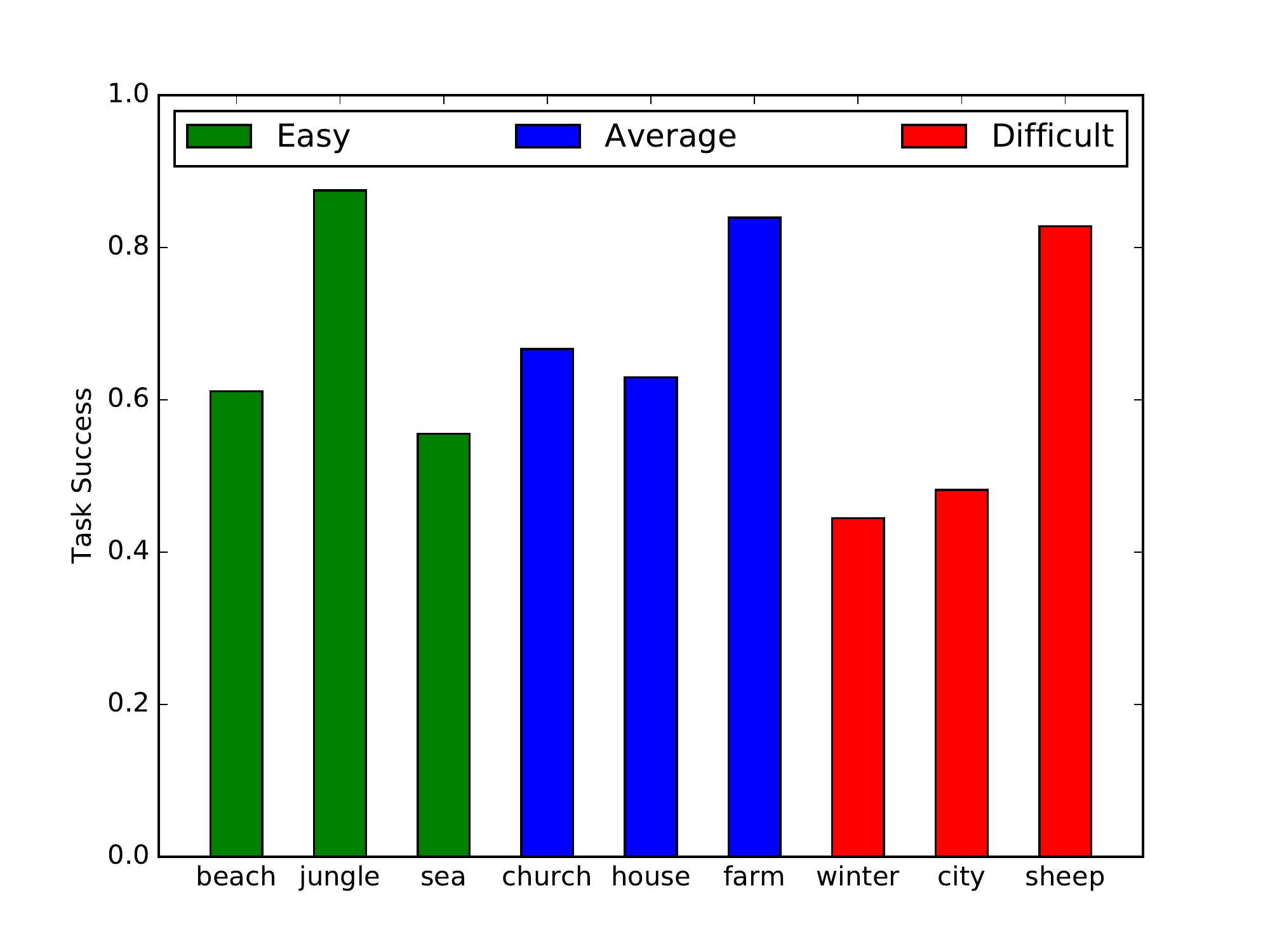}
\caption{Task success per scene and initially defined level of difficulty for each scene.}
\label{fig:taskSuccessscene}
\end{center}
\end{figure}

Another interesting result is the plot displayed in Figure \ref{fig:taskSuccessscene}. The initial division of the scenes according to the difficulty level does not seem to correspond to the average relative number of differences found. The scenes beach (Figure \ref{fig:picdiff}) and sea (Figure \ref{fig:sea}) seemed more complex than all the scenes in the average group and the sheep scene (Figure \ref{fig:sheep}) in the difficult group. 

\begin{figure*}[t]
    \centering
    \begin{subfigure}[t]{0.3\textwidth}
        \centering
        \includegraphics[width=\columnwidth]{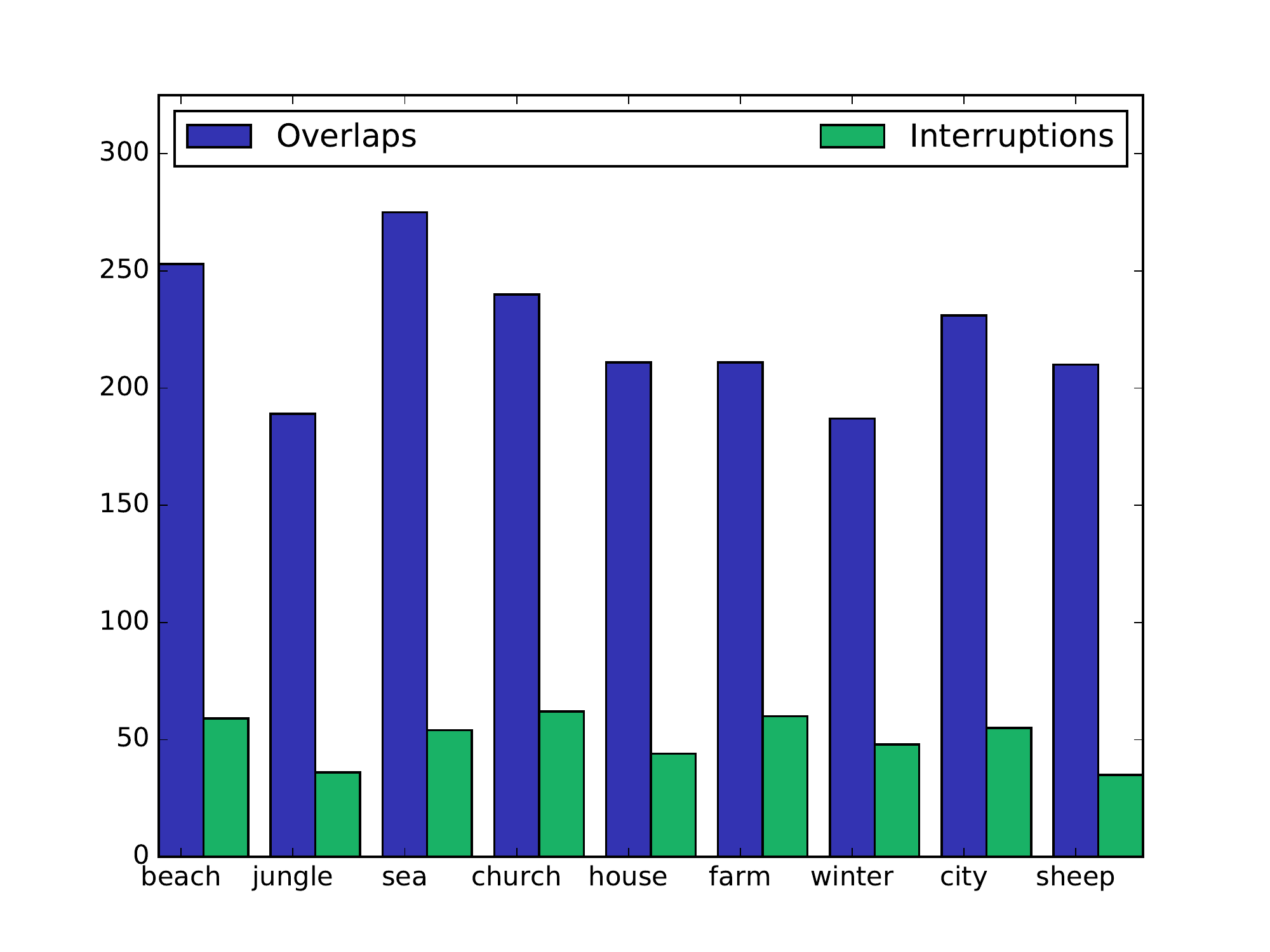}
        \caption{Overlaps.}
        \label{fig:overlapVsInt}
    \end{subfigure}%
    ~ 
    \begin{subfigure}[t]{0.3\textwidth}
        \centering
        \includegraphics[width=\columnwidth]{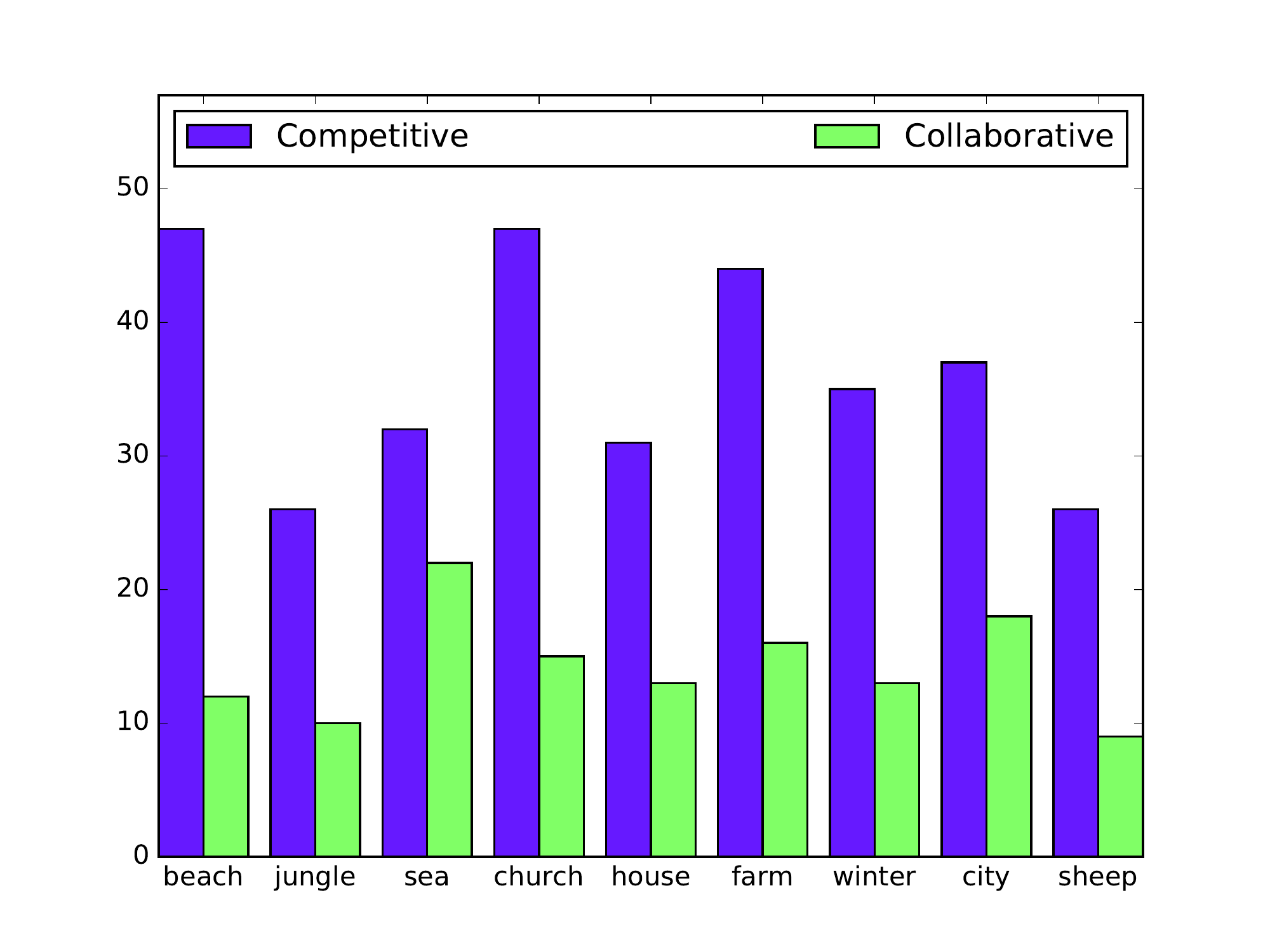}
        \caption{Interruptions.}
        \label{fig:collabVScomp}
    \end{subfigure}
    ~
       \begin{subfigure}[t]{0.3\textwidth}
        \centering
        \includegraphics[width=\columnwidth]{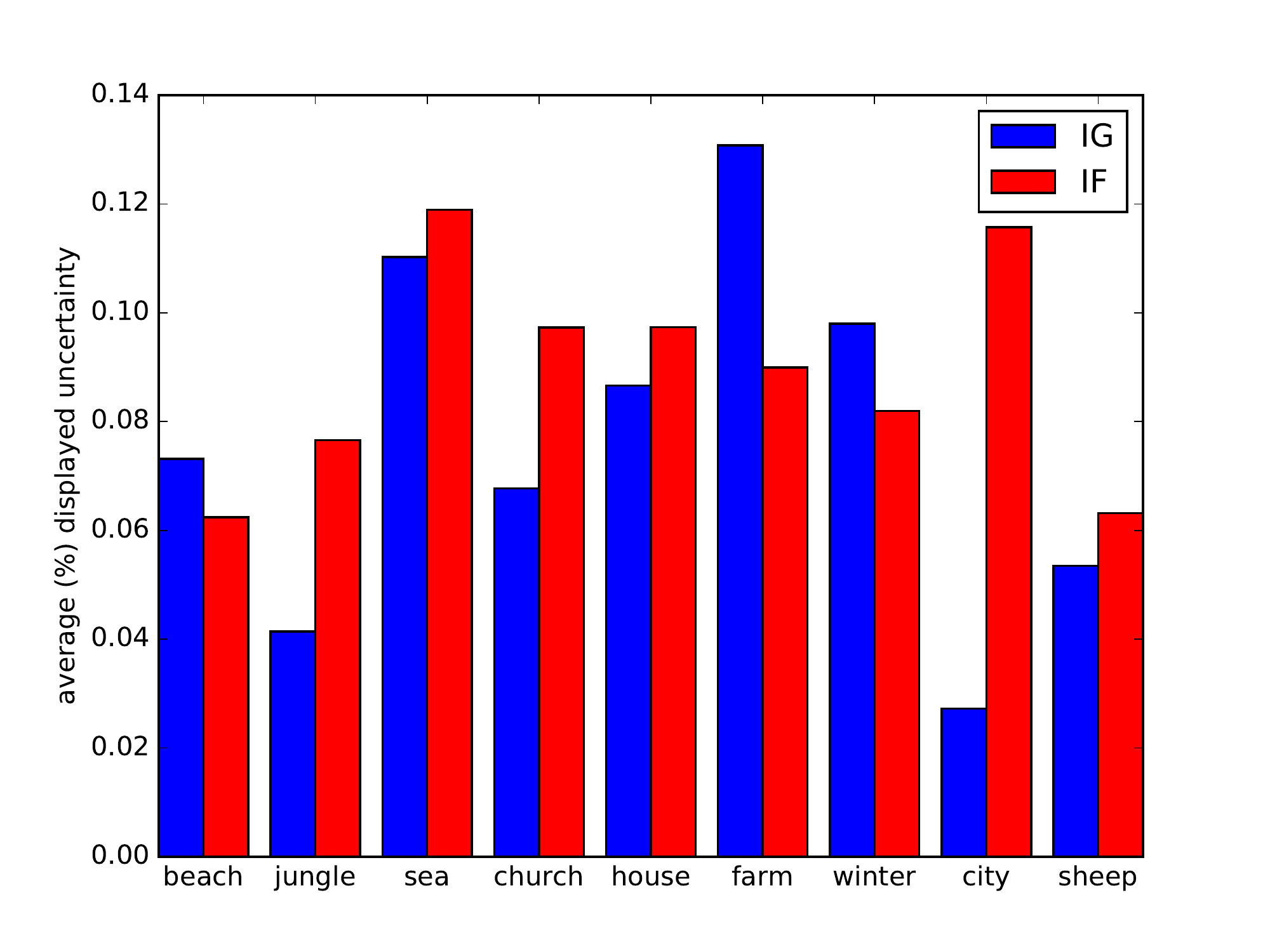}
        \caption{Uncertainty.}
        \label{fig:uncertainty}
    \end{subfigure}
    \caption{Overlaps (left), interruptions (center) and uncertainty (right) per scene.}
    \label{fig:breakdownFeatures}
\end{figure*}

%
%

These results seem to indicate that there are other factors that contribute to the scene complexity other than the number of differences between them. Therefore, we hypothesize that there could be a way to correlate the scene complexity with the task success. For this, first we have computed the entropy of the histogram of the RGB components. We found out that, except for house (Figure \ref{fig:house}), the scenes with highest success rate (jungle in Figure \ref{fig:jungle}, farm in Figure \ref{fig:farm} and sheep) were those with the lowest entropy values. Another factor that could have also contributed to the complexity of the scene is the number of objects in the picture. This information could be easily obtained parsing the XML files for each picture and counting the number of objects defined there. In this process we can even differentiate between group and single objects. Combining this information with the entropy of the scene, we can hypothesize that the complexity of the scene in this setup could be given by a combination between entropy and the number of objects. The house scene (Figure \ref{fig:house}) has low entropy, since there is very little color variation, but it has 21 objects (median value is 15). On the other hand, the scene sea (Figure \ref{fig:sea}) has 9 objects in the XML description, but the entropy value in $3.99$ (median value is $3.57$).

We hypothesize that the complexity of the scene would also have an impact on the mechanisms that mediate the interaction. Therefore we present a preliminary study where we tried to relate interruptions and uncertainty to the scene complexity. Just like in dancing, when the steps of the dance are more difficult, there is a higher chance that people step on each other's feet.
Figure \ref{fig:overlapVsInt} shows the comparison between the number the total of overlaps and the number of overlaps followed by an interruption, Figure \ref{fig:collabVScomp} shows the comparison between the number of competitive and collaborative interruptions per scene and Figure \ref{fig:uncertainty} shows the average time where participants showed uncertainty for each picture. The graph from Figures \ref{fig:overlapVsInt}, particularly concerning the number of interruptions per picture shows a similar trend to the graph show in Figure \ref{fig:taskSuccessscene}, that is the scenes with high average relative number of differences found were those with lower average number of interruptions, particularly competitive interruptions. 

Overlaps do not show a similar trend. In Figure \ref{fig:collabVScomp} similar trends hold, and the pictures with the lowest number of collaborative interruptions are exactly those where participants where the success rate was higher. The graph in Figure \ref{fig:uncertainty}, does not present the same trends a the previous two. We further explored interactions between uncertainty, interruptions and overlaps. We found a significant interaction between interruptions and uncertainty for both IGs, IFs and both participants combined in Chi-square test performed ($p-value<0.05$, $p-value<0.001$ and $p<0.01$, respectively). Regarding overlaps and uncertainty, we also found significant differences for IFs and both participants combined ($p-value<0.05$ and $p<0.001$, respectively), but not for IGs according to the Chi-square test performed. These are interesting results, which can indicate that a multi-modal approach for breakdown detection can be worth investigating. In dialogue like in dancing, if someone steps on the partners foot, this might have impacts in the rest of the movements the body needs to perform.

We made a further analysis regarding uncertainty to assess the difficulty of predicting uncertainty from facial features and gaze. Using OpenFace \cite{baltruvsaitis2016openface} to extract a set of facial features together with the variance of the gaze movement. After trying several different techniques we achieved an 62\% accuracy as our best result (54\% was be the majority baseline) using Artificial Neural Networks in an evenly distributed subset of the data (the original data set is highly skewed towards certain segments).




\section{Conclusion}

This paper presented the Spot the Difference corpus, a corpus of spontaneous task-oriented spoken dialogues. The set-up used, the experimental procedure and the participants data were described in detail, together with the annotations performed. Parts of data is publicly available online and the complete dataset can be obtain via the first author. It includes a meta-description of the data, audio, video, pictures and respective XML descriptions, the annotations described in this paper and code to parse the data logs.

As we have shown in our preliminary analyses, this data offers various possibilities regarding the study of mechanisms that regulate human communication. For instance, one could look at how humans ground different representation of very similar scenes and how it interacts with turn-taking.

To sum up, we think that this corpus contributes to future research in the multi-modal spoken interaction, so that soon we can have spoken dialogue systems which are able to dance with their users.

\section*{Acknowledgments}
The authors would like to thank Anna Hjalmarsson and Catharine Oertel for the discussions on data collection, annotation and analysis, Zofia Malisz for the help to with stats, Cl\'{a}udia Velhas for transcribing the data, Casey Kennington for the helpful comments and discussions, and reviewers for the sharp suggestions.

\section{Bibliographical References}
\label{main:ref}

\bibliographystyle{lrec}
\bibliography{lrec2018}

\begin{thebibliography}{}

\bibitem[\protect\citename{Anderson \bgroup et al.\egroup
  }1991]{anderson1991hcrc}
Anderson, A.~H., Bader, M., Bard, E.~G., Boyle, E., Doherty, G., Garrod, S.,
  Isard, S., Kowtko, J., McAllister, J., Miller, J., et~al.
\newblock (1991).
\newblock The hcrc map task corpus.
\newblock {\em Language and speech}, 34(4):351--366.

\bibitem[\protect\citename{Baltru{\v{s}}aitis \bgroup et al.\egroup
  }2016]{baltruvsaitis2016openface}
Baltru{\v{s}}aitis, T., Robinson, P., and Morency, L.-P.
\newblock (2016).
\newblock Openface: an open source facial behavior analysis toolkit.
\newblock In {\em Applications of Computer Vision (WACV), 2016 IEEE Winter
  Conference on}, pages 1--10. IEEE.

\bibitem[\protect\citename{Brennan and Clark}1996]{brennan1996conceptual}
Brennan, S.~E. and Clark, H.~H.
\newblock (1996).
\newblock Conceptual pacts and lexical choice in conversation.
\newblock {\em Journal of Experimental Psychology: Learning, Memory, and
  Cognition}, 22(6):1482.

\bibitem[\protect\citename{Bull and Aylett}1998]{bull1998analysis}
Bull, M. and Aylett, M.~P.
\newblock (1998).
\newblock An analysis of the timing of turn-taking in a corpus of goal-oriented
  dialogue.
\newblock In {\em ICSLP}.

\bibitem[\protect\citename{Colman and Healey}2011]{colman2011distribution}
Colman, M. and Healey, P.
\newblock (2011).
\newblock The distribution of repair in dialogue.
\newblock In {\em Proceedings of the Annual Meeting of the Cognitive Science
  Society}, volume~33.

\bibitem[\protect\citename{DeVault}2008]{devault2008contribution}
DeVault, D.
\newblock (2008).
\newblock {\em Contribution tracking: participating in task-oriented dialogue
  under uncertainty}.
\newblock Rutgers The State University of New Jersey-New Brunswick.

\bibitem[\protect\citename{Friedberg \bgroup et al.\egroup
  }2012]{friedberg2012lexical}
Friedberg, H., Litman, D., and Paletz, S.~B.
\newblock (2012).
\newblock Lexical entrainment and success in student engineering groups.
\newblock In {\em Spoken Language Technology Workshop (SLT), 2012 IEEE}, pages
  404--409. IEEE.

\bibitem[\protect\citename{Garrod and Anderson}1987]{GarrodAnderson87}
Garrod, S.~C. and Anderson, A.
\newblock (1987).
\newblock {S}aying what you mean in dialogue: a study in conceptual and
  semantic co-ordination.
\newblock {\em Cognition}, 27:181--218.

\bibitem[\protect\citename{Goldberg}1990]{goldberg1990interrupting}
Goldberg, J.~A.
\newblock (1990).
\newblock Interrupting the discourse on interruptions: An analysis in terms of
  relationally neutral, power-and rapport-oriented acts.
\newblock {\em Journal of Pragmatics}, 14(6):883--903.

\bibitem[\protect\citename{John \bgroup et al.\egroup }2008]{john2008paradigm}
John, O.~P., Naumann, L.~P., and Soto, C.~J.
\newblock (2008).
\newblock Paradigm shift to the integrative big five trait taxonomy.
\newblock {\em Handbook of personality: Theory and research}, 3:114--158.

\bibitem[\protect\citename{Kennington \bgroup et al.\egroup
  }2013]{kennington2013interpreting}
Kennington, C., Kousidis, S., and Schlangen, D.
\newblock (2013).
\newblock Interpreting situated dialogue utterances: an update model that uses
  speech, gaze, and gesture information.
\newblock In {\em Proceedings of the SIGDIAL 2013 Conference}, pages 173--182.

\bibitem[\protect\citename{Lai \bgroup et al.\egroup }2013]{lai2013summarize}
Lai, C., Carletta, J., and Renals, S.
\newblock (2013).
\newblock Detecting summarization hot spots in meetings using group level
  involvement and turn-taking features.
\newblock In {\em Proc. Interspeech 2013, Lyon, France}.

\bibitem[\protect\citename{Liscombe \bgroup et al.\egroup
  }2005]{liscombe2005detecting}
Liscombe, J., Hirschberg, J., and Venditti, J.~J.
\newblock (2005).
\newblock Detecting certainness in spoken tutorial dialogues.
\newblock In {\em Interspeech 2005}.

\bibitem[\protect\citename{Moniz \bgroup et al.\egroup
  }2014]{moniz2014speaking}
Moniz, H., Batista, F., Mata, A.~I., and Trancoso, I.
\newblock (2014).
\newblock Speaking style effects in the production of disfluencies.
\newblock {\em Speech Communication}, 65:20--35.

\bibitem[\protect\citename{Nenkova \bgroup et al.\egroup
  }2008]{nenkova2008high}
Nenkova, A., Gravano, A., and Hirschberg, J.
\newblock (2008).
\newblock High frequency word entrainment in spoken dialogue.
\newblock In {\em Proceedings of the 46th annual meeting of the association for
  computational linguistics on human language technologies: Short papers},
  pages 169--172. Association for Computational Linguistics.

\bibitem[\protect\citename{Oviatt \bgroup et al.\egroup
  }2015]{oviatt2015spoken}
Oviatt, S., Hang, K., Zhou, J., and Chen, F.
\newblock (2015).
\newblock Spoken interruptions signal productive problem solving and domain
  expertise in mathematics.
\newblock In {\em Proceedings of the 2015 ACM on International Conference on
  Multimodal Interaction}, pages 311--318. ACM.

\bibitem[\protect\citename{Poesio and Rieser}2010]{poesio2010completions}
Poesio, M. and Rieser, H.
\newblock (2010).
\newblock Completions, coordination, and alignment in dialogue.
\newblock {\em Dialogue and Discourse}, 1(1):1--89.

\bibitem[\protect\citename{Reitter and Moore}2007]{reitter2007predicting}
Reitter, D. and Moore, J.~D.
\newblock (2007).
\newblock Predicting success in dialogue.
\newblock In {\em Proceedings of the 45th Annual Meeting of the Association of
  Computational Linguistics}, page 808{\textendash}815, Prague, Czech Republic,
  June.

\bibitem[\protect\citename{Schrank and Schuppler}2015]{schrank2015automatic}
Schrank, T. and Schuppler, B.
\newblock (2015).
\newblock Automatic detection of uncertainty in spontaneous german dialogue.
\newblock In {\em Sixteenth Annual Conference of the International Speech
  Communication Association (Interspeech)}.

\bibitem[\protect\citename{Skantze and {Al Moubayed}}2012]{Skantze2012}
Skantze, G. and {Al Moubayed}, S.
\newblock (2012).
\newblock Iristk: a statechart-based toolkit for multi-party face-to-face
  interaction.
\newblock In {\em Proceedings of ICMI}.

\bibitem[\protect\citename{Zarrie{\ss} \bgroup et al.\egroup
  }2016]{zarriess2016pentoref}
Zarrie{\ss}, S., Hough, J., Kennington, C., Manuvinakurike, R., DeVault, D.,
  Fernandez, R., and Schlangen, D.
\newblock (2016).
\newblock Pentoref: A corpus of spoken references in task-oriented dialogues.
\newblock In {\em 10th edition of the Language Resources and Evaluation
  Conference}.

\end{thebibliography}

\newpage

\begin{appendices}
\section{Scenes}

\begin{figure}[H]
\centering
\includegraphics[width=0.6\columnwidth]{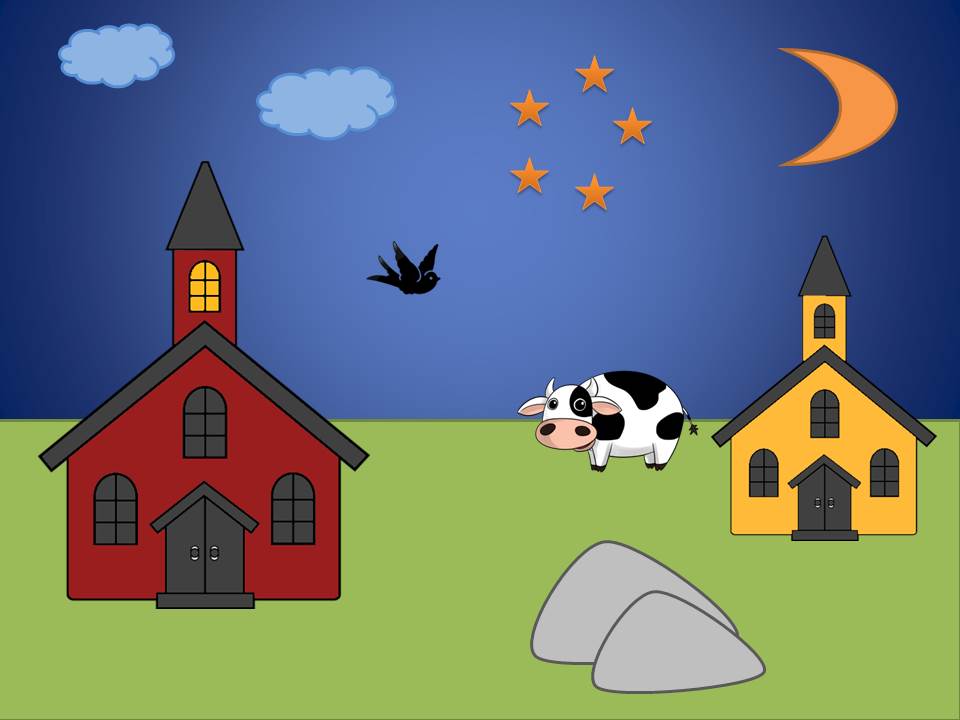}
\caption{Church scene.}
\label{fig:church}
\end{figure}

\begin{figure}[H]
\centering
\includegraphics[width=0.6\columnwidth]{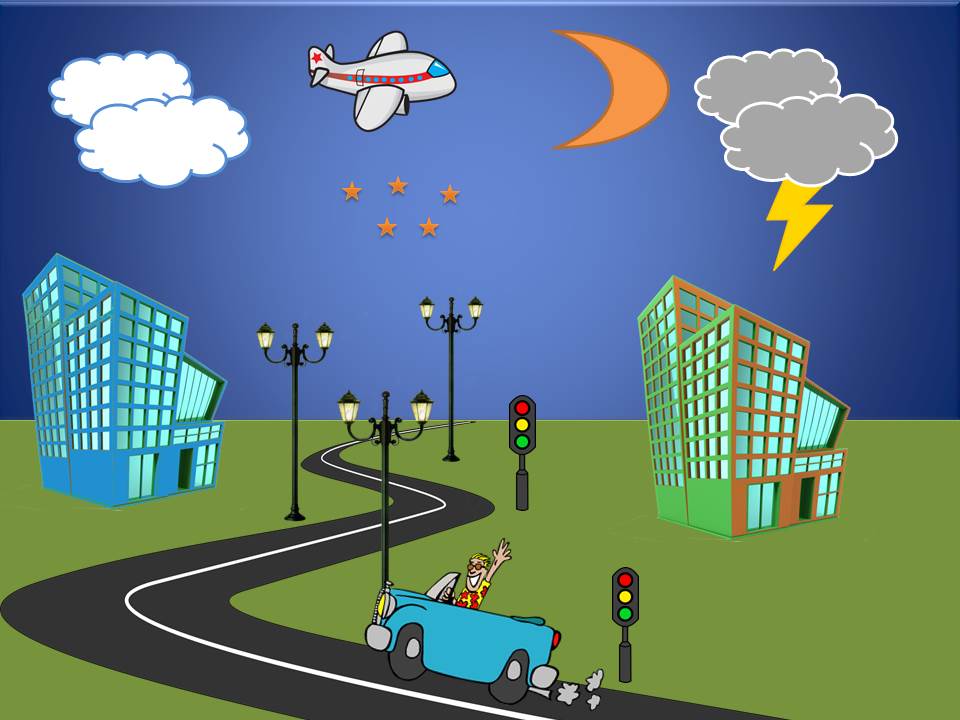}
\caption{City scene.}
\label{fig:city}
\end{figure}

\begin{figure}[H]
\centering
\includegraphics[width=0.6\columnwidth]{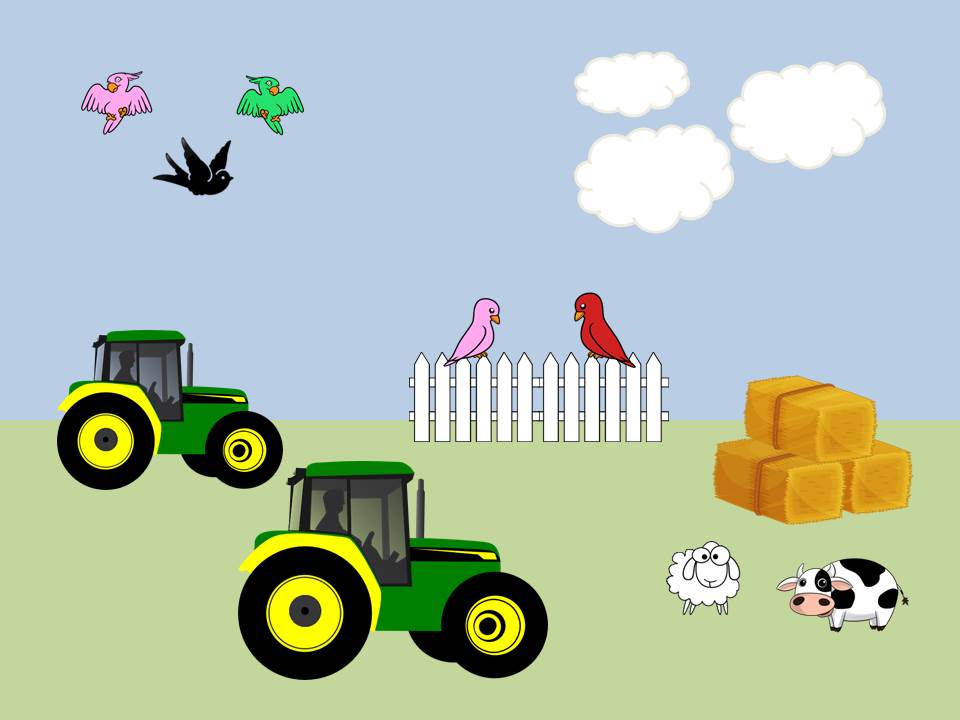}
\caption{Farm scene.}
\label{fig:farm}
\end{figure}

\begin{figure}[H]
\centering
\includegraphics[width=0.6\columnwidth]{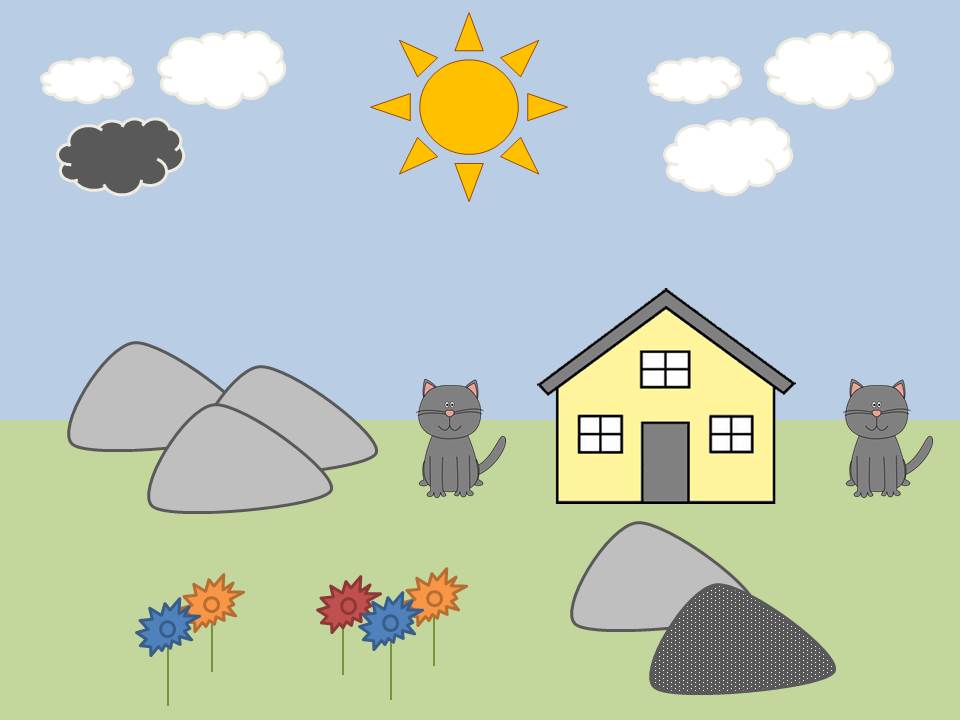}
\caption{House scene.}
\label{fig:house}
\end{figure}	

\begin{figure}[H]
\centering
\includegraphics[width=0.6\columnwidth]{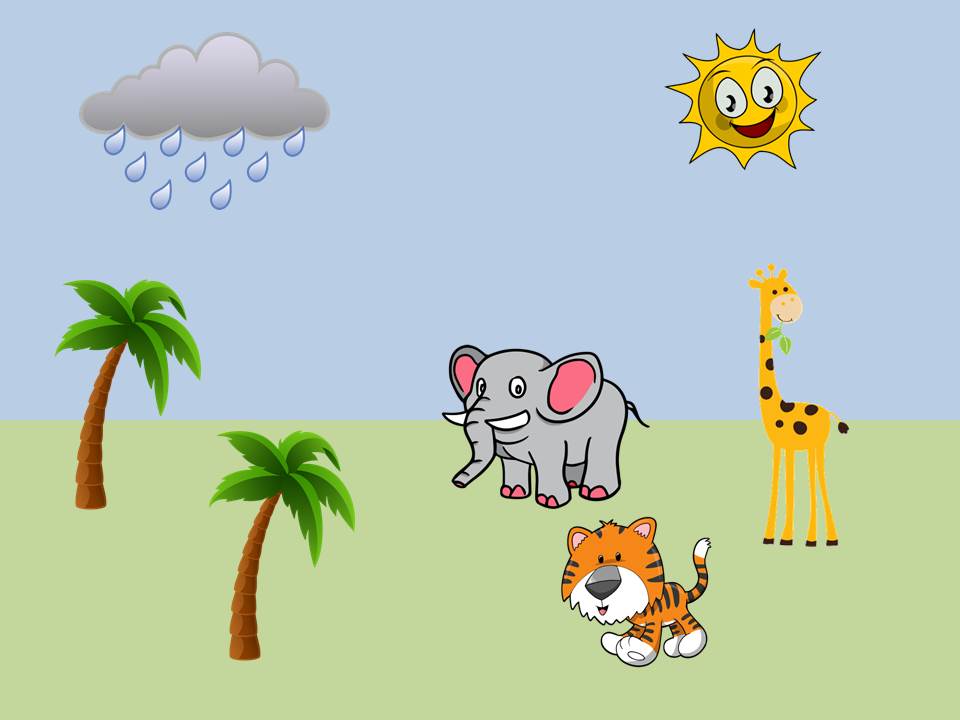}
\caption{Jungle scene.}
\label{fig:jungle}
\end{figure}

\begin{figure}[H]
\centering
\includegraphics[width=0.6\columnwidth]{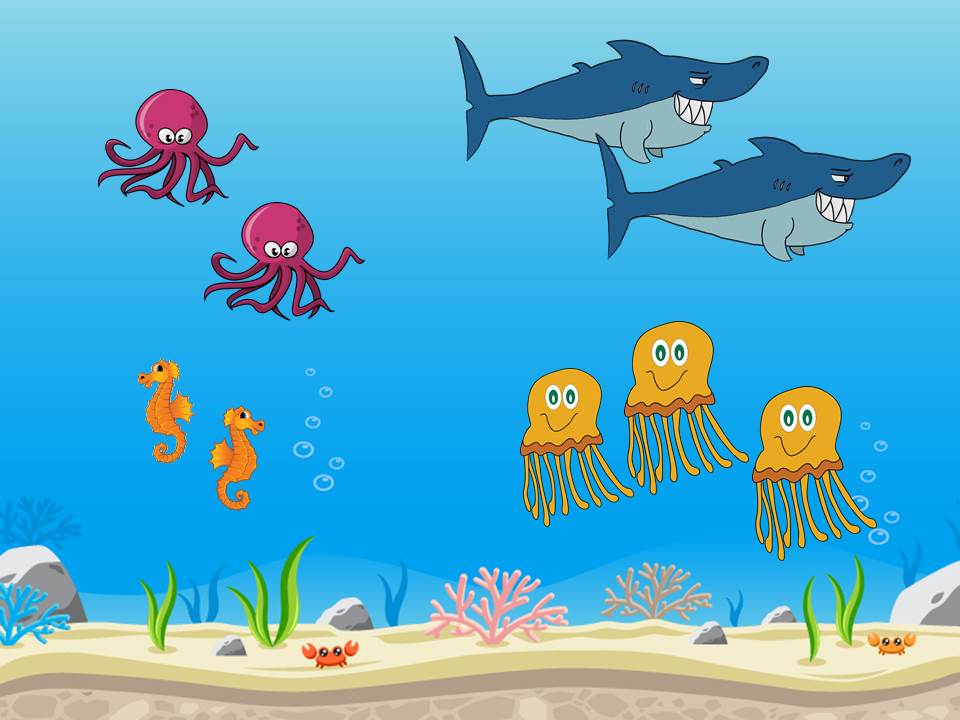}
\caption{Sea scene.}
\label{fig:sea}
\end{figure}		

\begin{figure}[H]
\centering
\includegraphics[width=0.6\columnwidth]{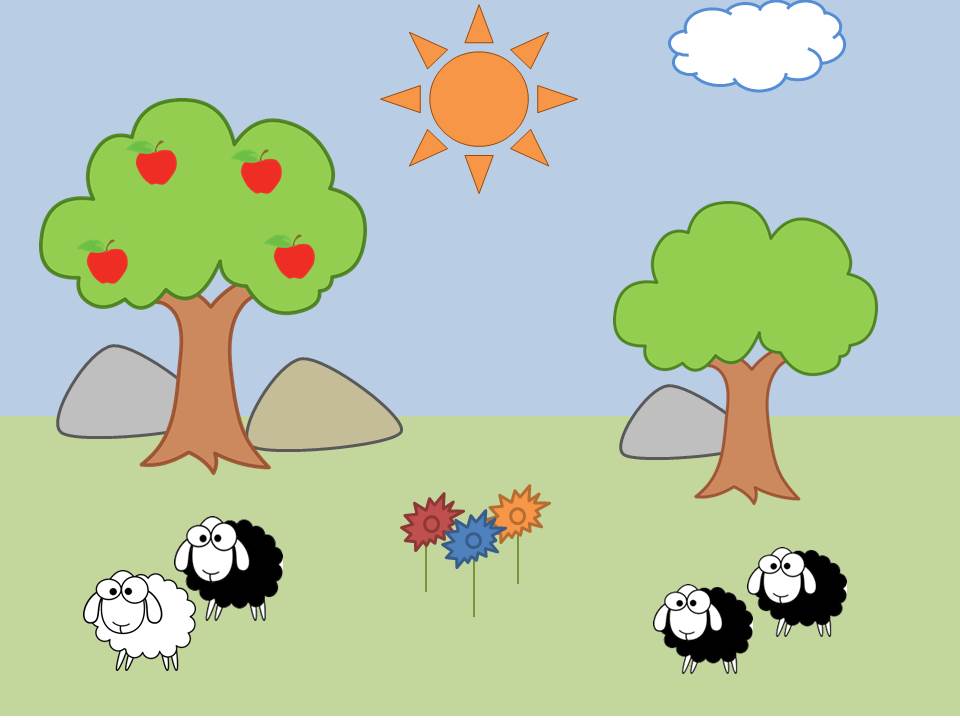}
\caption{Sheep scene.}
\label{fig:sheep}
\end{figure}

\begin{figure}[H]
\centering
\includegraphics[width=0.6\columnwidth]{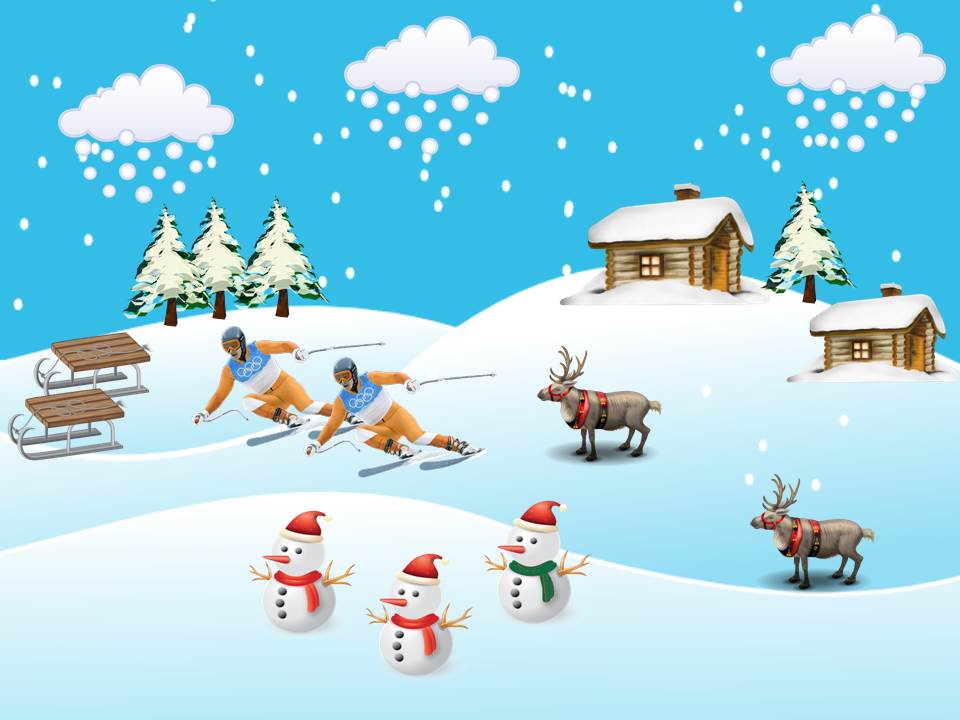}
\caption{Winter scene.}
\label{fig:winter}
\end{figure}

\end{appendices}

\end{document}